\documentclass[10pt,conference]{IEEEtran}
\IEEEoverridecommandlockouts
\usepackage{cite}
\usepackage{amsmath,amssymb,amsfonts}
\usepackage{algorithmic}
\usepackage{graphicx}
\usepackage{textcomp}
\usepackage[colorlinks,linkcolor=red,anchorcolor=green,citecolor=blue]{hyperref}
\usepackage{xcolor}
\def\BibTeX{{\rm B\kern-.05em{\sc i\kern-.025em b}\kern-.08em
    T\kern-.1667em\lower.7ex\hbox{E}\kern-.125emX}}
\usepackage{booktabs}
\usepackage[ruled]{algorithm2e}  
\usepackage{amsmath,amssymb,amsfonts,bm}
\usepackage{multirow}
\newcommand{\hl}{\textcolor{blue}}

\usepackage{hyperref}

\usepackage{authblk}

\begin{document}

\title{Optimizing Memory Efficiency of Graph Neural Networks on Edge Computing Platforms}
\IEEEspecialpapernotice{[ Brief Industry Paper ]
\thanks{This work is supported by National Natural Science Foundation of China (Grant No. 62072019, 61602022). Corresponding author is Jianlei Yang, Email: \url{jianlei@buaa.edu.cn}. The source code of this paper is publicly available on:
\url{https://github.com/BUAA-CI-Lab/GNN-Feature-Decomposition}.}}
\author[1,2]{Ao Zhou}
\author[2]{Jianlei Yang}
\author[2]{Yeqi Gao}
\author[2]{Tong Qiao}
\author[2]{Yingjie Qi}
\author[1]{Xiaoyi Wang}
\author[1]{Yunli Chen}
\author[3]{\\Pengcheng Dai}
\author[4]{Weisheng Zhao}
\author[2]{Chunming Hu}

\affil[1]{School of Software, Beijing University of Technology, Beijing, China}
\affil[2]{School of Computer Science and Engineering, Beihang University, Beijing, China}

\affil[3]{Beijing Bytedance Technology Co., Ltd, Beijing, China}
\affil[4]{School of Integrated Circuit Science and Engineering, Beihang University, Beijing, China}
\renewcommand\Authands{ and }


\maketitle

\begin{abstract}
Graph neural networks (GNN) have achieved state-of-the-art performance on various industrial tasks.
However, the poor efficiency of GNN inference and frequent Out-Of-Memory (OOM) problem limit the successful application of GNN on edge computing platforms.
To tackle these problems, a feature decomposition approach is proposed for memory efficiency optimization of GNN inference.
The proposed approach could achieve outstanding optimization on various GNN models, covering a wide range of datasets, which speeds up the inference by up to $3\times$.
Furthermore, the proposed feature decomposition could significantly reduce the peak memory usage (up to $5\times$ in memory efficiency improvement) and mitigate OOM problems during GNN inference.
\end{abstract}




\section{Introduction}
In recent years, as a generalization of conventional deep learning methods on the non-Euclidean domain, Graph Neural Networks (GNN) are widely applied to various fields of research, such as node classification \cite{kipf2016semi}, link prediction \cite{zhang2018link} and feature matching \cite{sarlin2020superglue}.
In this paper, we propose a memory efficient method - feature decomposition, for GNN inference on hardware resource-limited platforms (mobiles, edge devices, etc.).
We focus on two major problems in GNN inference: 1) \textit{Poor inference efficiency} - The irregular graph structure and large vertex feature length pose difficulties in efficient GNN inference on edge devices,
and 2) \textit{Frequent Out-Of-Memory (OOM) problem} - The feature vectors with high dimensionality occupy enormous space in memory that often exceeds the memory limit.

%

The Message Passing based GNN inference can be summarized as two distinct phases: Combination and Aggregation.
The former updates feature vector of each vertex with MLP operations, while the latter updates the vectors by aggregating features in their neighborhoods.
The majority of current works on GNN optimization focus on large-scale high performance systems with abundant resources (GPU/Memory/etc.) \cite{Jia2020,jia2020improving}. 
Unfortunately, there exists little research explored for improving GNN inference efficiency on CPU-only edge devices.
In this paper, we propose a novel approach to optimize GNN memory efficiency from a new perspective: decomposing the dimension of feature vectors and performing aggregation respectively.
Considering that aggregate operation is an element-wise operation on feature vectors, decomposing the feature vectors of all vertices and performing aggregation separately will not harm the accuracy of GNN inference. 
With our method, the data reuse of neighbor feature vectors is improved by loading more feature data into cache, which greatly improves aggregation efficiency.
Besides, the greatly reduced feature dimension in each aggregate operation also reduces the risk of encountering OOM.

To evaluate our feature decomposition method, we conduct sufficient experiments with PyG framework \cite{Fey/Lenssen/2019}.
Our feature decomposition approach can be easily implemented based on Gather-ApplyEdge-Scatter (GAS) abstraction of PyG.
The proposed approach is evaluated for different GNN inference models with various datasets and could obtain about $3\times$ speedups compared with PyG baseline.
Furthermore, our approach could significantly improve the memory efficiency by reducing the cache miss rate and peak memory usage. 

\section{Feature Decomposition}

We first profile GNNs performance on CPU to figure out their workload characteristics, and then describe the proposed memory optimization approaches in detail.

\subsection{Characterizing GNNs on CPU}
To identify the computation bottleneck of GNNs on resource-limited CPU, we conduct quantitative characterizations using PyG on Intel CPU. 
Three classic GNN models are chosen as evaluation objects, including GCN \cite{kipf2016semi}, GraphSage (GSC) \cite{Hamilton2017} and GAT \cite{velivckovic2017graph}. 
\begin{figure}[t]
    \centering
    \includegraphics[width = 0.95\linewidth]{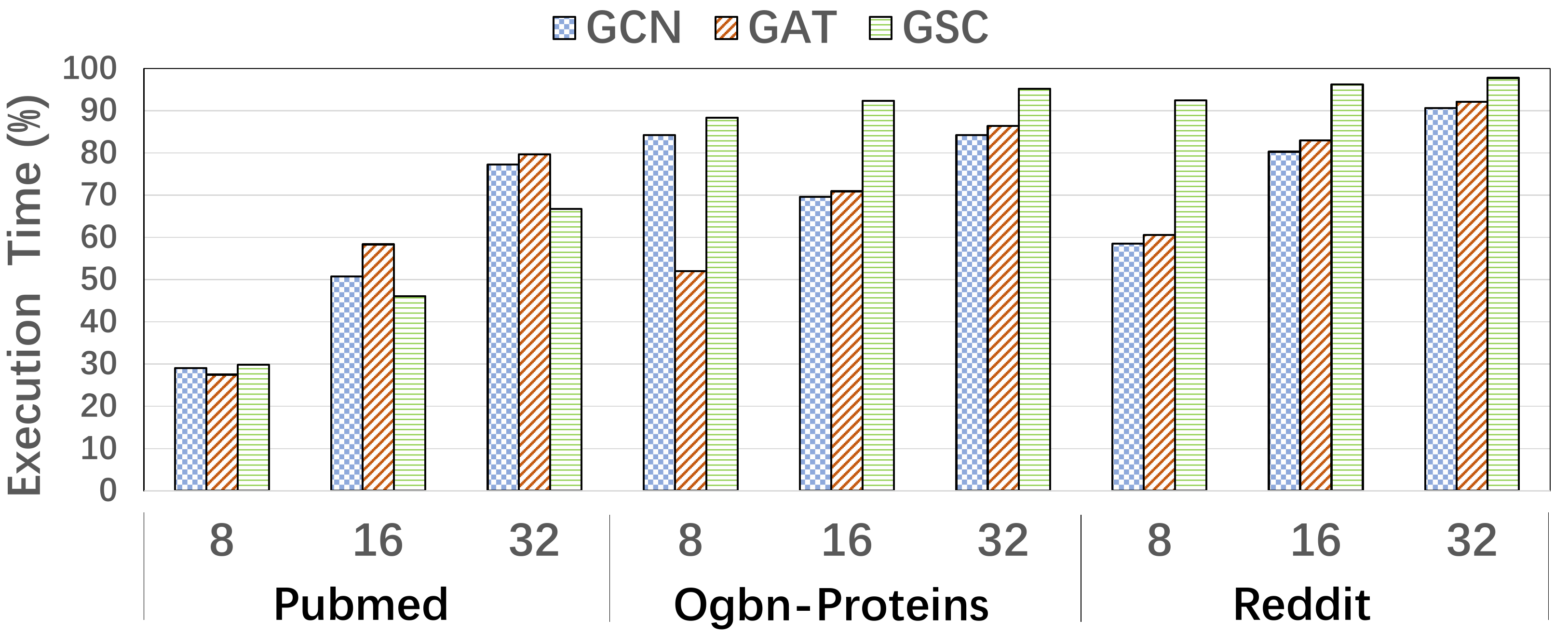}
    \caption{Percentage of aggregation runtime comparison for GNNs inference (with hidden dimension of $8$, $16$ and $32$). }
    \label{fig:execution}
\end{figure}

\textbf{Execution Time Breakdown}. We evaluate GNNs inference on several datasets  \cite{sen2008collective,hu2020ogb,Hamilton2017},
and the percentage of aggregation runtime comparison is illustrated in Fig. \ref{fig:execution}. 
Due to the large scale of graph and long feature vectors, the aggregation phase usually dominates the performance of whole inference procedure.
Especially for Reddit dataset, aggregation phase occupies above $97\%$ of the whole execution time. 
Even for the small-scale Pubmed dataset, it still takes up more than half of the inference computation cost if the hidden dimension (output dimension of combination phase) exceeds $16$.
\begin{table}[t]   
    \centering
    \caption{GNNs Characterization on CPU.}   
    \begin{tabular}{l r r}    
        \toprule    &{Aggregation} &{Combination}      \\    
        \midrule  \textbf{Exeucted IPC (IPC)}        &  $0.46$    &  $0.87$     \\  
                  \textbf{L1 Cache Miss Rate (L1 Miss)}  &  $41.86\%$ &  $7.54\%$   \\   
                  \textbf{LL Cache Miss Rate (LLC Miss)}  &  $45.59\%$ &  $34.05\%$  \\   
                  \textbf{TLB Cache Miss Rate (TLB Miss)} &  $9.18\%$  &  $0.15\%$   \\ 
        \midrule  \textbf{Data Reusability}    &  Low     &  High     \\  
                  \textbf{Execution Bound}     &  Memory  &  Computation  \\   
        \bottomrule 
    \end{tabular}
    \label{tab:cachemiss}
\end{table}

\textbf{Memory Access}. Table \ref{tab:cachemiss} summarizes the execution patterns for GCN on Reddit, with the hidden dimension of $32$.
It is observed that \textbf{L1 Cache Miss Rate} and \textbf{Last Level Cache Miss Rate} in the aggregation phase are extremely high.
Besides, the \textbf{Executed IPC} is only $0.46$. 
The low efficiency of cache utilization in the aggregation phase is caused by the high randomness of memory accessing and poor data reuse between neighbor vertex.
Only a few feature vectors can be loaded into the cache at the same time due to the high vertex feature dimension.
Additionally, the peak memory usage during inference is approximately equal to 32 GB, resulting in a relatively high risk of OOM.
Consequently, the aggregation phase of GNN is memory-bound, with irregular data access pattern and low data reusability.

\subsection{Proposed Approach}

\begin{figure}[t]
    \centering
    \includegraphics[width = 0.95\linewidth]{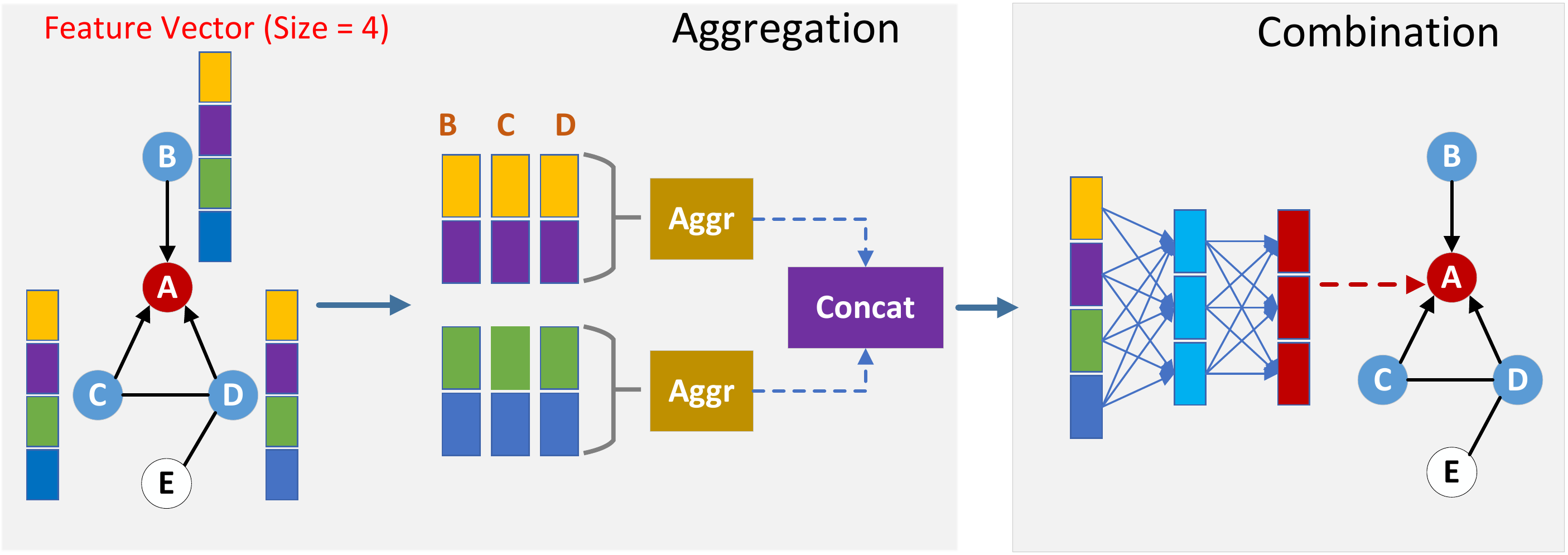}
    \caption{Computation of one vertex (in red) in a GNN layer by feature decomposition method. The number of decomposition layers $\cal P$ is $2$.}
    \label{fig:cut}
\end{figure}
To solve the above problems, we propose a new optimization method for GNN aggregation: \textbf{Feature Decomposition}, which can speed up GNN inference and avoid frequent OOM problem.
Since the high dimension of feature data increases the reuse distance between vertices and causes OOM problems during aggregation phase, we decompose the dimension into smaller ones to alleviate these problems.
In our method, the feature vector of each vertex is divided into $\cal P$ layers, and then the aggregation is performed layer by layer on all vertices. 
We take the computation of one vertex in a GNN layer as an example to describe our method. 

As shown in Fig. \ref{fig:cut}, vertex \textcircled{A} (in red) in the graph has three neighbors: \textcircled{B}, \textcircled{C}, \textcircled{D}. 
General GNN computation on vertex \textcircled{A} can be summarized into two steps.
Firstly, vertex \textcircled{A} aggregates its neighbors' feature vectors which are marked with blue, 
and then compute new activations of the vertex by MLP.
Unlike the general computation process, feature decomposition aims to decompose the feature vector before conducting the aggregation. 
In Fig. \ref{fig:cut}, we split the feature vector of all vertices into $2$ layers, perform aggregation on each layer respectively, and finally concatenate the feature vectors together.
After that, combination phase is performed to update the feature vector of vertex \textcircled{A}. 

\begin{algorithm}[t]
\small
\KwIn{Graph $G(V,E)$, Layers $\cal K$, vertex features $\bm{h}^{(0)}_v$ with length $L$ and the number of partitions $\cal P$. }
\KwOut{Updated vertex representations $\bm{z}_v$.}
\For{$k=1\dots \cal K$}{
    \hl{/* Feature vector decomposition for all vertices */}
 
    \For{$v\in V$}{ 
        $\bm{h}^{(k-1)}_v[{\cal{P}}]\leftarrow Chunk \left(
        \bm{h}^{(k-1)}_v,{\cal{P}},\dim=-1\right)$ 

    }
    \hl{/* Aggregation on each decomposition layer */}

    \For{$p\in \cal P$}{
        \For{$v\in V$}{
            $\bm{m}^{(k)}_{uv}[p]\leftarrow Message \left( {\bm{h}^{(k-1)}_u[p],\bm{h}^{(k-1)}_v[p]} \right)$
            
            $\bm{a}^{(k)}_v[p]\leftarrow Aggregate \left( \bm{m}^{(k)}_{uv}[p]\mid u\in N(v)  \right) $
        }
    }
    {$\bm{a}^{(k)}_v\leftarrow Concat \left(\bm{a}^{(k)}_v[0:p-1])\right)$
    $\bm{h}^{(k)}_v[p]\leftarrow Update \left(\bm{a}^{(k)}_v, \bm{h}^{(k-1)}_v \right)$
    }
}
$\bm{z}_v \leftarrow \bm{h}_v^{(\cal K)}$
\caption{GNN Inference Procedure with Feature Decomposition}
\label{gnnalgo}
\end{algorithm}

Algorithm \ref{gnnalgo} describes the complete computation procedure of GNN using our feature decomposition method. 
Same as the single node computation shown in Fig. \ref{fig:cut}, the feature vectors of all vertices are decomposed into $\cal P$ layers and aggregation is executed on these layers separately.
In fact, even though feature decomposition brings efficiency improvements by executing aggregation separately, it usually still brings some additional overhead.
%
Performing aggregation on each layer of feature vector requires repeated access to all edges.
Therefore, the number of decomposition layers should be determined according to the memory capacity of the device and characteristics of datasets (edges, feature dimension, etc.).

\subsection{Implementations with PyG Framework}

Feature decomposition is implemented with PyG framework by slight modifications.
PyG defines GNN execution paradigm based on {\texttt{MessagePassing}}.
The {\texttt{MessagePassing}} interface of PyG relies on a gather-scatter scheme to aggregate messages from neighboring vertices. 
Therefore, the input of the {\texttt{forward}} process is the edge set of graph in COO format and the feature vectors of all vertices.
The aggregation process of GNN is performed by {\texttt{propagate}}.
To provide users with a feature decomposition choice, we modified the {\texttt{forward}} function in the {\texttt{MessagePassing}} class. 
The specific modifications are mainly listed as follows:
\begin{itemize}
    \item A new parameter is added to the {\texttt{forward}} function: \texttt{Layers}. It means how many layers the feature vector is divided into.
    \item In the {\texttt{forward}} function, the input feature vector is divided by \texttt{torch.chunk} referring to \texttt{Layers}, and the {\texttt{propagate}} is performed respectively on all layers.
    \item After the {\texttt{propagate}} is completed, the feature vectors of each layer is concatenated together by \texttt{torch.concat}.
\end{itemize}


\section{Experimental Results}
\label{sec:exp}

The proposed feature decomposition approach is evaluated and compared with two aggregation schemes in PyG on Intel CPU and Raspberry Pi.
The evaluated graphs information are listed in Table \ref{tab:dataset},
where Pubmed \cite{sen2008collective} has fewer vertices and low average degree of vertex, Ogbn-Proteins \cite{hu2020ogb} and Reddit \cite{Hamilton2017} have larger data scale and higher average degree.
The granularity of feature decomposition embodies in the dimension of each decomposition layer, which is set as \{GCN: 1, GAT: 8, GSC: 4\} for Reddit and Ogbn-Proteins, and \{GCN: 4, GAT: 8, GSC: 4\} for Pubmed.
For GNN models, we select GCN \cite{kipf2016semi}, GSC \cite{Hamilton2017} and GAT \cite{velivckovic2017graph} to cover GNNs with different aggregate operators.
Among the two schemes in PyG, the Memory Efficient Aggregation scheme PyG\_MEA does not support the implementation of GAT, therefore, the corresponding data is omitted.

\begin{table}[t]
    \centering
    \caption{\label{tab:dataset}Information of evaluated graph datasets.}   
    \begin{tabular}{l r r r}    
        \toprule   Name   & \# Vertices  & \# Edges   & Avg. Degree \\   
        \midrule  
        Pubmed   & $19,717$  & $88,676$  & $4.5$ \\
        Ogbn-Proteins & $132,534$  & $39,561,252$ & $597$  \\
        Reddit & $232,965$  & $114,615,892$ & $492$  \\
        \bottomrule 
    \end{tabular} 
\end{table}

\begin{figure}[t]
    \centering
    \includegraphics[width = 1.0\linewidth]{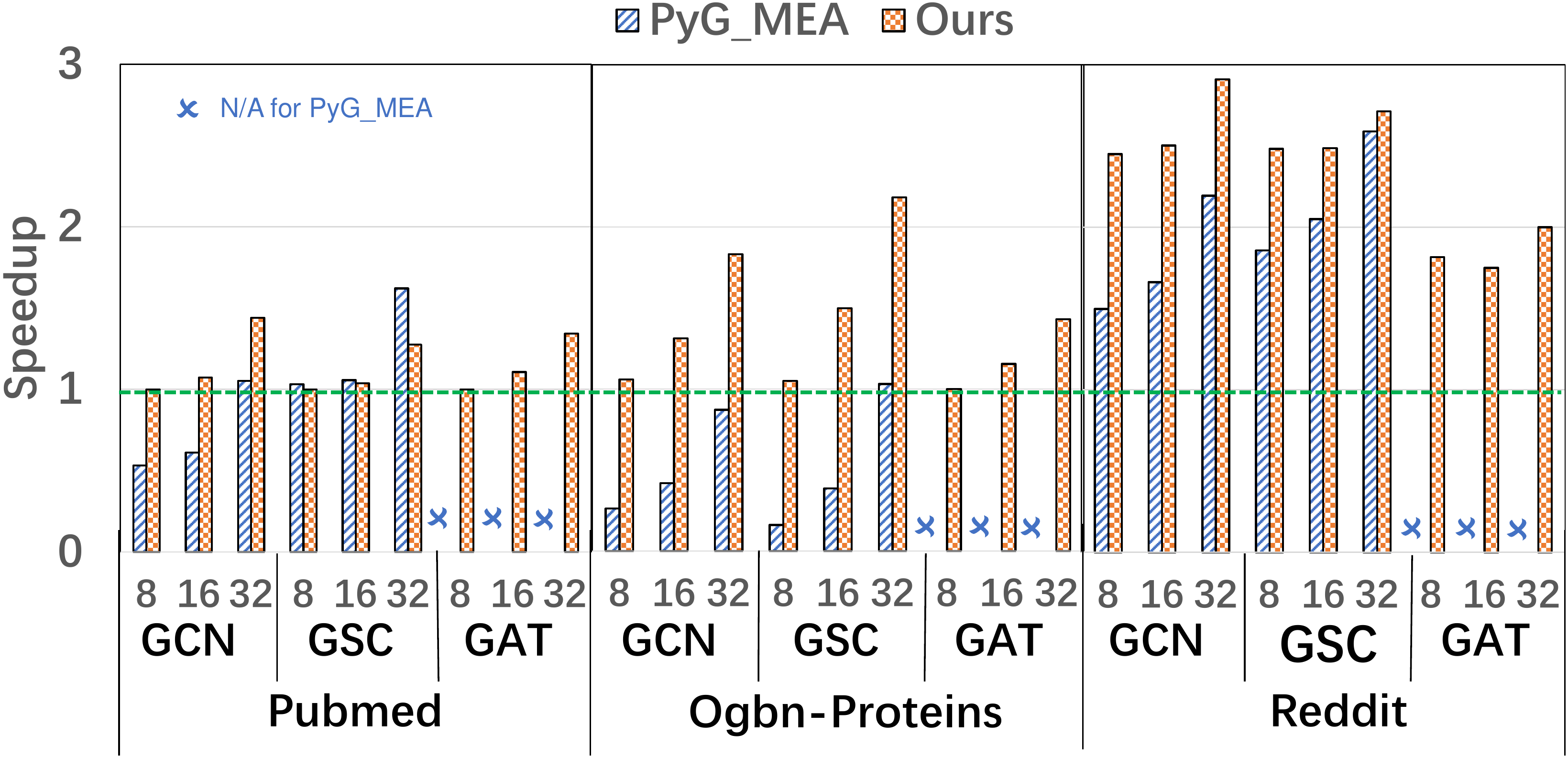}
    \caption{Performance comparison between our proposed approach and PyG. The PyG with standard aggregation scheme is marked by the green dash line as a baseline. The PyG\_MEA could obtain some speedups in Reddit dataset than baseline. Our approach could obtain significant speedups in many cases.}
    \label{fig:latency}
\end{figure}

\begin{figure}[t]
    \centering
    \includegraphics[width = 0.95\linewidth]{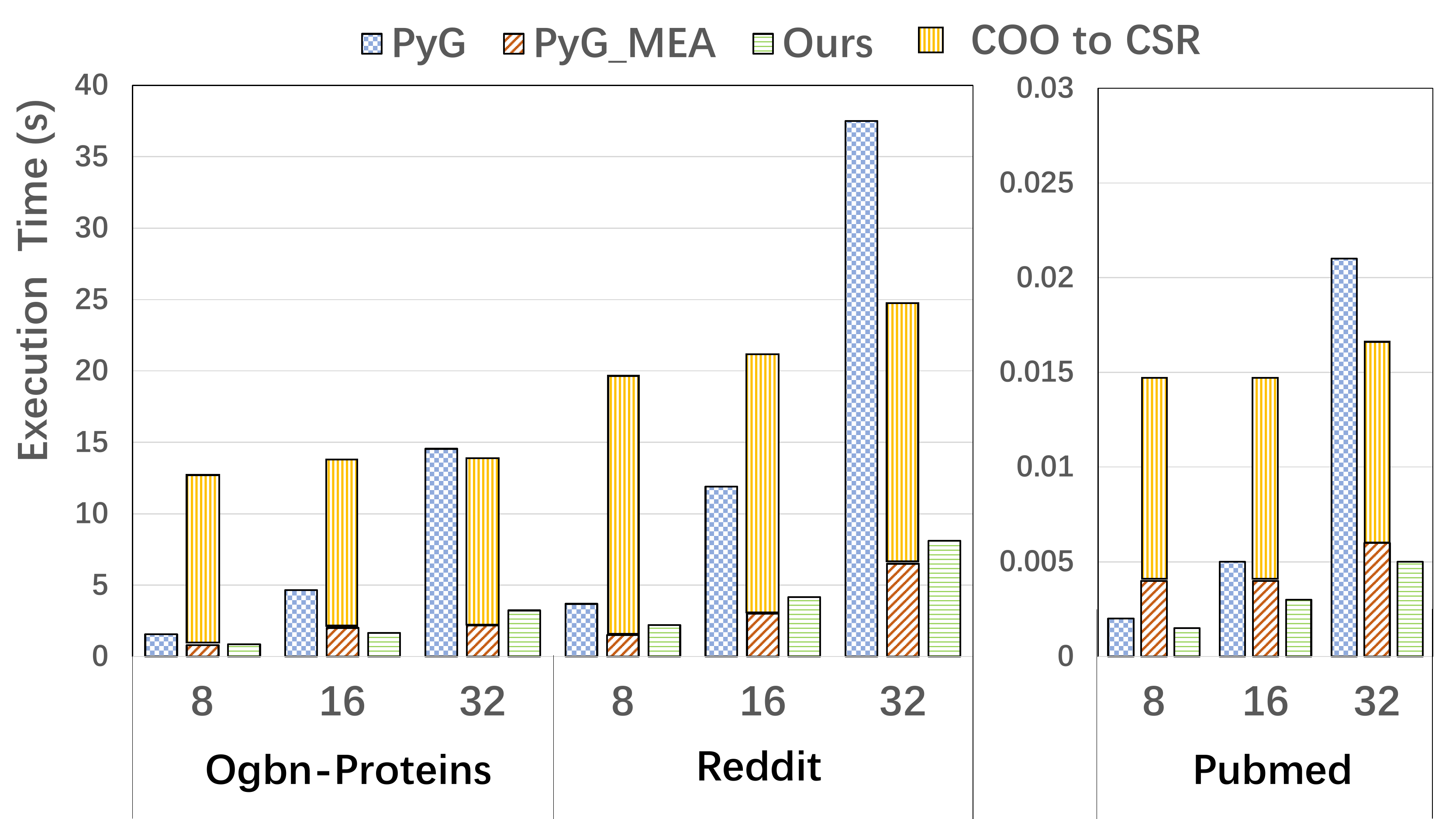}
    \caption{Latency of aggregation and format conversion.}
    \label{fig:cootocsr}
\end{figure}

Learn from previous experience \cite{yan2020characterizing}, we perform the combination phase ahead of aggregation, which usually helps to improve aggregation efficiency. 
As a result, the feature dimension in aggregation phase is determined by hidden dimension of combination. 
When the hidden dimension is set above $32$, the peak memory usage during GNN inference exceeds the limit of $32$ GB, thus causing OOM problem for all baseline methods.
Therefore, we set the hidden dimension as \{$8$,$16$,$32$\} for comparison between different methods, while also implementing our method with higher hidden dimension for further evaluation.

\subsection{Inference Latency}
As our first motivation is to reduce latency in GNN inference, here we first evaluate the latency of our feature decomposition method across several benchmarks.
The relative end-to-end speedups over PyG under different configurations are summarized in Fig. \ref{fig:latency}.
The results show that the speedup provided by our method is significant (up to $3\times$) and universal, with benefits on large graphs and high average degree.

PyG\_MEA does not perform well on datasets with high degree and short lengh of feature vectors due to excessive format conversion overhead.
As shown in Fig. \ref{fig:cootocsr}, PyG\_MEA takes nearly $13$ seconds to perform format conversion on Ogbn-Proteins.
The overhead cannot be offset by the improvement of its aggregation efficiency.
In contrast, our method has no data conversion overhead and is more efficient.

\begin{figure}[t]
    \centering
    \includegraphics[width = 0.9\linewidth]{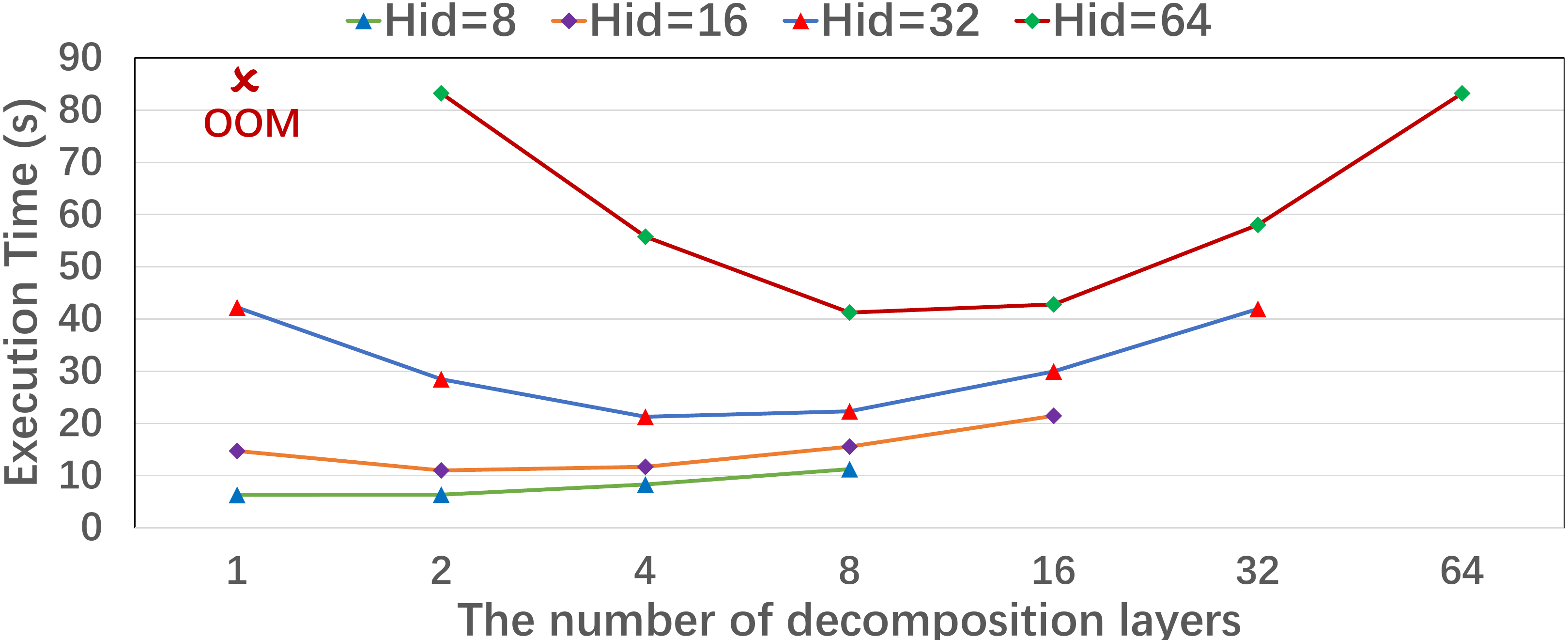}
    \caption{Feature decomposition evaluation of GAT on Reddit with various decomposition layers and hidden dimensions. Hid means hidden dimension. $\times$ OOM means out-of-memory appeared when with Hid=$64$ and only $1$ decomposition layer deployed.}
    \label{fig:layer}
\end{figure}

\subsection{Impact of Feature Decomposition Granularity}
As we described earlier, feature decomposition has a tradeoff between feature data reusability and edge data utilization.
%
%
Different decomposition layers are evaluated with various hidden dimensions as shown in Fig. \ref{fig:layer}. 
As the number of decomposition layers increases, the optimization brought by feature data reuse will be gradually neutralized with redundant memory access of edge data.
Therefore, the number of decomposition layers should be selected reasonably to obtain the optimal efficiency in aggregation.
Furthermore, with the expansion of the hidden dimension, our method can achieve better speedup.
In addition, as long as the aggregation phase can be executed when hidden dimension is $1$, our method can avoid the OOM problem.
As shown in Fig. \ref{fig:layer}, the baselines have OOM problem with hidden dimension $64$.
In contrast, GAT optimized by our method can still inference efficiently.

\subsection{Memory Efficiency}

Memory efficiency is evaluated by analyzing cache miss rate and peak memory usage. Comparison results between our approach and PyG are illustrated in Table \ref{tab:test}.

\textbf{Cache Access Performance}.
Our method could significantly reduce the cache miss rate and improve executed IPC, because the reuse of neighbor feature data has been improved by reducing the feature vector length during aggregation phase.

\begin{table}[t]   
    \centering
    \caption{Profiling results of GCN/GAT/GSC on Reddit with Hid=$32$.}   
    \begin{tabular}{c c r r r r}    
        \toprule  
        Model & Method & {L1 Miss} &{LLC Miss} &{TLB Miss} &{IPC}  \\ 
        \midrule  
        \multirow{2}{*}{GCN} &  PyG  & $23.15\%$   &  $43.49\%$    &  $4.43\%$    & $0.66$  \\ 
                  ~          &  Ours & \textbf{6.94\%}  &  \textbf{5.09\%}   &  \textbf{0.05\%}  & \textbf{1.82} \\
        \midrule
        \multirow{2}{*}{GAT} &  PyG  & $24.18\%$   &  $43.01\%$    &  $4.58\%$    & $0.63$  \\ 
                  ~          &  Ours & \textbf{10.39\%}   &  \textbf{9.75\%} &  \textbf{0.44\%}  & \textbf{1.44} \\ 
        \midrule
        \multirow{2}{*}{GSC} &  PyG  & $40.47\%$   &  $42.27\%$    &  $8.69\%$    & $0.48$  \\  
                  ~          &  Ours & \textbf{15.71\%}  &  \textbf{3.48\%}  &  \textbf{0.09\%} & \textbf{1.25} \\  
        \bottomrule
    \end{tabular} 
    \label{tab:test}
\end{table} 

\begin{figure}[t]
    \centering
    \includegraphics[width = 0.8\linewidth]{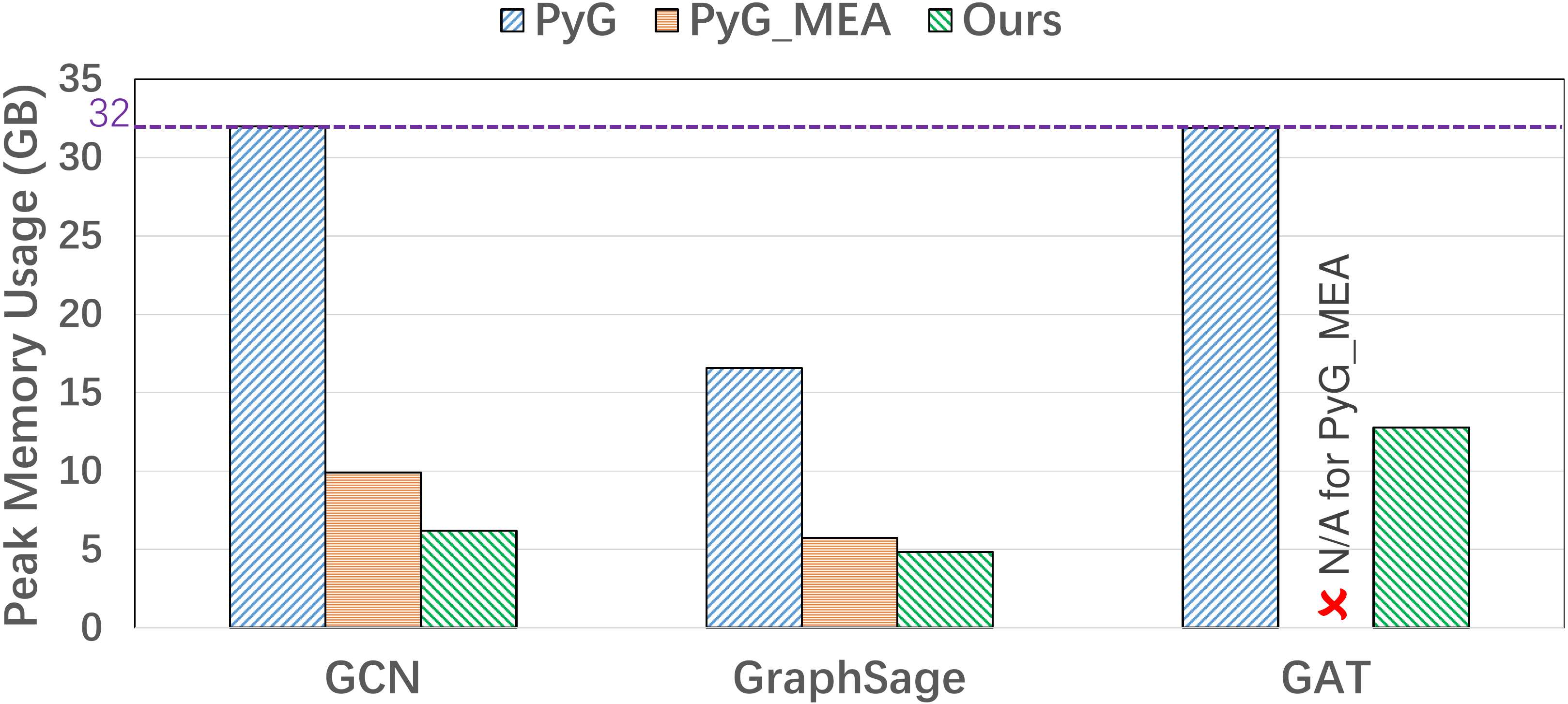}
    \caption{Peak memory usage comparison for the first layer of GCN/GAT/GSC evaluated on Reddit, with hidden dimension Hid=$32$. PyG\_MEA is not available for running GAT.}
    \label{fig:memory}
\end{figure}

\textbf{Peak Memory Usage}.
Fig. \ref{fig:memory} compares the peak memory usage of different aggregation schemes with hidden dimension Hid=$32$.
It is obvious that our method can significantly reduce the peak memory usage, i.e., the memory efficiency is improved up to $5\times$. 
The GCN inference on Reddit with our method only requires $6$ GB memory, which effectively alleviates the frequent OOM problem when performing GNN inference on the large-scale graphs.

\subsection{Evaluations on Raspberry Pi Device}
\begin{figure}[t]
    \centering
    \includegraphics[width = 0.95\linewidth]{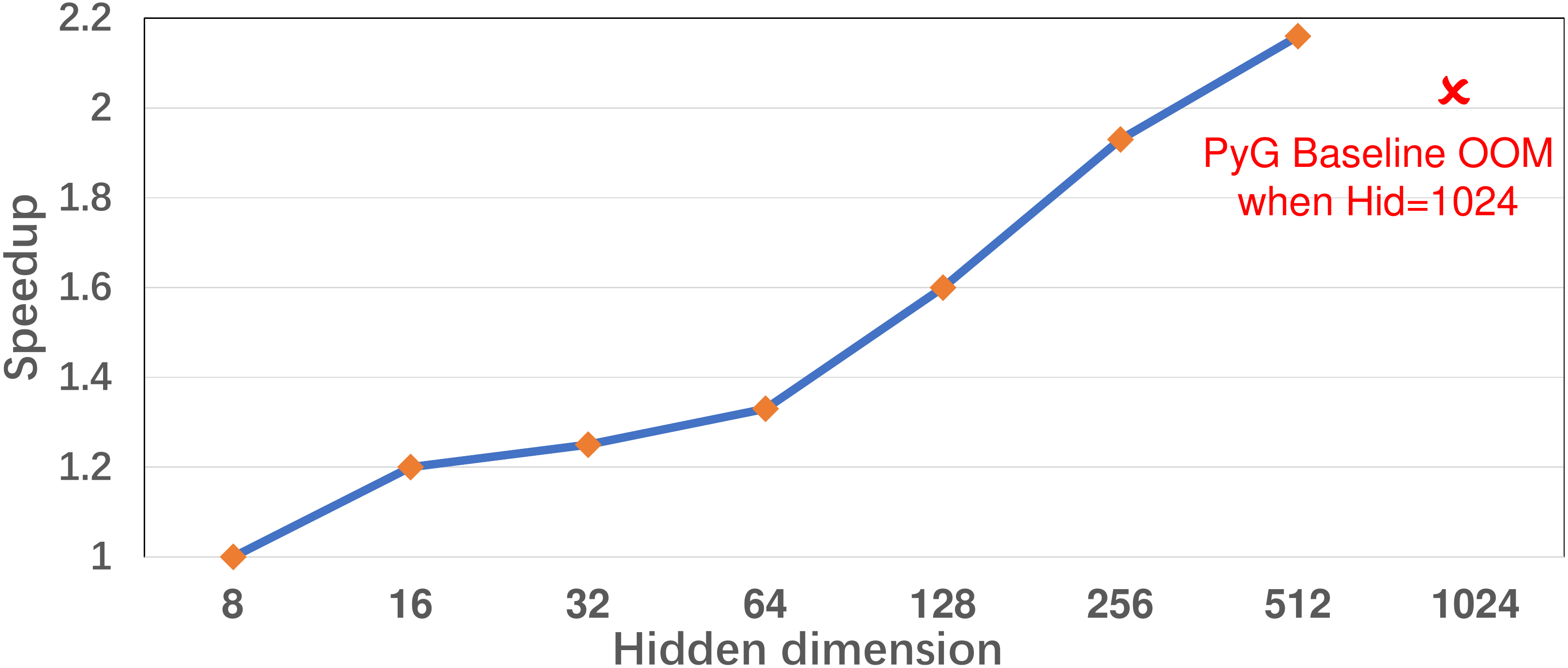}
    \caption{Speedup of GCN with our method on Raspberry Pi compared with original PyG, where PyG has OOM when hidden dimension exceeds $1024$.}
    \label{fig:raspberry}
\end{figure}

To further evaluate our method on resource-limited edge devices, we chose Raspberry Pi with only $1$ GB memory as the experimental platform.
Due to the extremely limited hardware resources, only Pubmed is evaluated and the speedups are shown in Fig. \ref{fig:raspberry}. As the hidden dimensions increasing, we could achieve better inference performance. For a very large hidden dimension, Hid=$1024$, PyG will has OOM while our approach could still works efficiently.

\section{Conclusions}
This work proposes feature decomposition for optimizing memory efficiency of GNN inference. Especially for the OOM problems in edge devices due to the very limited available hardware resources, it could significantly reduce the required peak memory.
The results have shown that our method can provide a great boost for GNN edge computing applications.

\bibliographystyle{unsrt}
\begingroup
\bibliography{main}

\begin{thebibliography}{10}

\bibitem{kipf2016semi}
Thomas~N Kipf and Max Welling.
\newblock Semi-supervised classification with graph convolutional networks.
\newblock {\em arXiv preprint arXiv:1609.02907}, 2016.

\bibitem{zhang2018link}
Muhan Zhang and Yixin Chen.
\newblock Link prediction based on graph neural networks.
\newblock In {\em Proceedings of NIPS}, pages 5165--5175, 2018.

\bibitem{sarlin2020superglue}
Paul-Edouard Sarlin, Daniel DeTone, Tomasz Malisiewicz, and Andrew Rabinovich.
\newblock Superglue: Learning feature matching with graph neural networks.
\newblock In {\em Proceedings of CVPR}, pages 4938--4947, 2020.

\bibitem{Jia2020}
Zhihao Jia, Sina Lin, Rex Ying, Jiaxuan You, Jure Leskovec, and Alex Aiken.
\newblock Redundancy-free computation for graph neural networks.
\newblock In {\em Proceedings of SIGKDD}, page 997–1005, 2020.

\bibitem{jia2020improving}
Zhihao Jia, Sina Lin, Mingyu Gao, Matei Zaharia, and Alex Aiken.
\newblock Improving the accuracy, scalability, and performance of graph neural
  networks with roc.
\newblock {\em Machine Learning and Systems}, 2:187--198, 2020.

\bibitem{Fey/Lenssen/2019}
Matthias Fey and Jan~E. Lenssen.
\newblock Fast graph representation learning with {PyTorch Geometric}.
\newblock In {\em Proceedings of ICLR}, 2019.

\bibitem{Hamilton2017}
William~L. Hamilton, Rex Ying, and Jure Leskovec.
\newblock Inductive representation learning on large graphs.
\newblock In {\em Proceedings of NIPS}, page 1025–1035, 2017.

\bibitem{velivckovic2017graph}
Petar Veli{\v{c}}kovi{\'c}, Guillem Cucurull, Arantxa Casanova, Adriana Romero,
  Pietro Lio, and Yoshua Bengio.
\newblock Graph attention networks.
\newblock {\em arXiv preprint arXiv:1710.10903}, 2017.

\bibitem{sen2008collective}
Prithviraj Sen, Galileo Namata, Mustafa Bilgic, Lise Getoor, Brian Galligher,
  and Tina Eliassi-Rad.
\newblock Collective classification in network data.
\newblock {\em AI Magazine}, 29(3):93--93, 2008.

\bibitem{hu2020ogb}
Weihua Hu, Matthias Fey, Marinka Zitnik, Yuxiao Dong, Hongyu Ren, Bowen Liu,
  Michele Catasta, and Jure Leskovec.
\newblock Open graph benchmark: Datasets for machine learning on graphs.
\newblock {\em arXiv preprint arXiv:2005.00687}, 2020.

\bibitem{yan2020characterizing}
Mingyu Yan, Zhaodong Chen, Lei Deng, Xiaochun Ye, Zhimin Zhang, Dongrui Fan,
  and Yuan Xie.
\newblock Characterizing and understanding {GCNs} on {GPU}.
\newblock {\em IEEE Computer Architecture Letters}, 19(1):22--25, 2020.

\end{thebibliography}
\endgroup
\vspace{12pt}
\color{red}
\end{document}